\documentclass[10pt,twocolumn,letterpaper]{article}

\usepackage{cvpr}              %

\usepackage{colortbl}
\usepackage{graphicx}
\usepackage{amsmath}
\usepackage{amssymb}
\usepackage{color}
\usepackage{booktabs}
\usepackage{enumitem}
\usepackage{tabularx} 
\usepackage{arydshln}
\usepackage{cuted}
\usepackage[table]{xcolor}
\definecolor{newlightblue}{RGB}{0,75,255}
\usepackage[pagebackref,breaklinks,colorlinks,urlcolor={newlightblue}, citecolor={newlightblue}]{hyperref}
\usepackage{array}
\usepackage[outline]{contour}
\usepackage{multirow}
\usepackage{xspace}
\usepackage[font=small]{caption}
\newcommand{\projecturl}{https://hellomuffin.github.io/exif-as-language}

\usepackage{graphicx, amsmath, amssymb, caption, subcaption, multirow, overpic, textpos}
\usepackage[british, english, american]{babel}

\newlength\savewidth

\newcolumntype{x}[1]{>{\centering\arraybackslash}p{#1pt}}
\newcolumntype{y}[1]{>{\raggedright\arraybackslash}p{#1pt}}
\newcolumntype{z}[1]{>{\raggedleft\arraybackslash}p{#1pt}}

\newcommand{\app}{\raise.17ex\hbox{$\scriptstyle\sim$}}

\definecolor{deemph}{gray}{0.6}

\definecolor{baselinecolor}{gray}{.9}
\newcommand{\baseline}[1]{\cellcolor{baselinecolor}{#1}}

\definecolor{lightyellow}{RGB}{255,255,170}
\definecolor{lightgray}{RGB}{240,240,240}

\newcommand{\mypar}[1]{\vspace{-3mm}\paragraph{#1}}
\newcommand{\fig}[1]{Fig.~\ref{#1}}

\newcommand{\tbl}[1]{Table~\ref{#1}}

\newcommand{\bv}[0]{\mathbf v}
\newcommand{\bm}[0]{\mathbf m}

\newcommand{\pmAP}[0]{\texttt{\contourlength{0.2pt}\contournumber{10}\contour{black}{p-mAP}}}
\newcommand{\cIoU}[0]{\texttt{\contourlength{0.2pt}\contournumber{10}\contour{black}{cIoU}}}

\newcommand{\resize}[0]{\small \contourlength{0.1pt}\contournumber{10}\contour{black}{resize}}
\newcommand{\crop}[0]{\small \contourlength{0.1pt}\contournumber{10}\contour{black}{crop}}

\newcommand{\quotes}[1]{``{#1}''}

\usepackage[pagebackref,breaklinks,colorlinks]{hyperref}

\hypersetup{
    colorlinks=true,
    urlcolor=magenta,
    }

\usepackage[capitalize]{cleveref}
\crefname{section}{Sec.}{Secs.}
\Crefname{section}{Section}{Sections}
\Crefname{table}{Table}{Tables}
\crefname{table}{Tab.}{Tabs.}

\def\cvprPaperID{5172} %
\def\confName{CVPR}
\def\confYear{2023}

\begin{document}

\title{EXIF as Language: Learning  Cross-Modal \\ Associations Between Images and Camera Metadata}
\author{Chenhao Zheng
\qquad
Ayush Shrivastava 
\qquad
Andrew Owens \vspace{1mm} \\
University of Michigan \vspace{1mm} \\
{\small {\href{\projecturl}{\projecturl}}}
}

\maketitle

\begin{strip}
\centering
    \centering
    \raggedright
    \vspace{-8mm} %
\includegraphics[width=\textwidth]{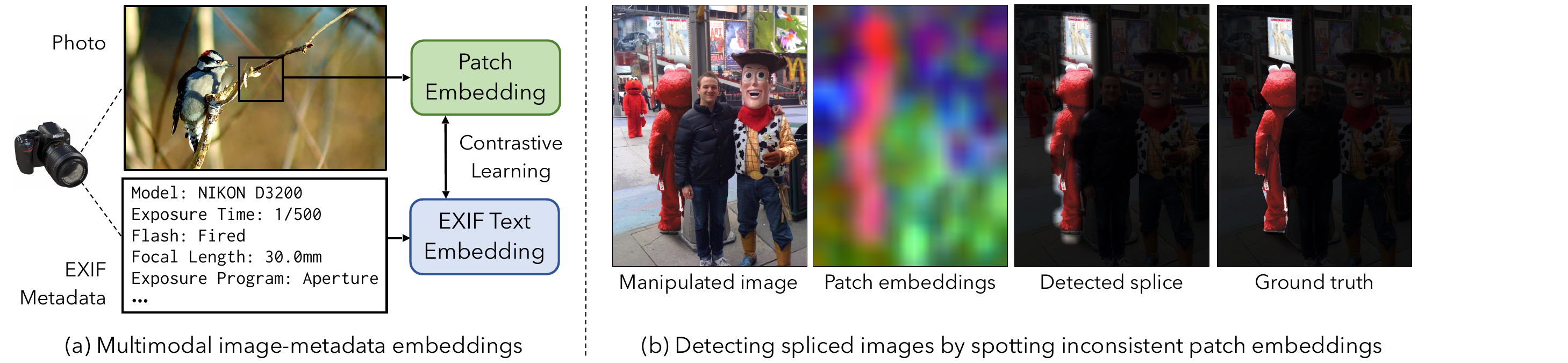}
\vspace{-4mm}
    \captionof{figure}{(a) We learn a joint embedding between image patches and the EXIF metadata that cameras automatically insert into image files. 
  Our model treats this metadata as a language-like modality: we convert the EXIF tags to text, concatenate them together, and then processes the result with a transformer.
      (b) We apply our representation to tasks that require understanding camera properties. For example, we can detect image splicing \quotes{zero shot} (and without metadata at test time) by finding inconsistent embeddings within an image. We show a manipulated image that contains content from two source photos. Since these photos were captured with different cameras, the two regions have dissimilar embeddings (visualized by PCA). We localize the splice by clustering the image's patch embeddings.}
  
  \vspace{0mm}
\label{fig:teaser}
\end{strip}

\begin{abstract}

We learn a visual representation that captures information about the camera that recorded a given photo. To do this, we train a multimodal embedding between image patches and the EXIF metadata that cameras automatically insert into image files. Our model represents this metadata by simply converting it to text and then processing it with a transformer. The features that we learn significantly outperform other self-supervised and supervised features on  downstream image forensics and calibration tasks. In particular, we successfully localize spliced image regions ``zero shot'' by clustering the visual embeddings for all of the patches within an image.

\end{abstract}

\section{Introduction}
\looseness=-1
A major goal of the computer vision community has been to use cross-modal associations to learn concepts that would be hard to glean from images alone \cite{multimodal}.
A particular focus has been on learning high level semantics, such as objects, from other rich sensory signals, like language and sound \cite{owens2018audio, CLIP}.
By design, the representations learned by these approaches typically discard {\em imaging} properties, such as the type of camera that shot the photo, its lens, and the exposure settings, which are not useful for their cross-modal prediction tasks~\cite{context_prediction}.

We argue that obtaining a complete understanding of an image requires both capabilities --- for our models to perceive not only the semantic content of a scene, but also the properties of the camera that captured it. This type of low level understanding has proven crucial for a variety of tasks, from image forensics~\cite{huh2018fighting, mantraNet, noi} to 3D reconstruction \cite{3d_construction,3d_construction_2}, yet it has not typically been a focus of representation learning.  It is also widely used in image generation, such as when users of text-to-image tools specify camera properties with phrases like ``DSLR photo''~\cite{ramesh2021zero,dalleprompt2022}.

We propose to learn low level imaging properties from the abundantly available (but often neglected) {\em camera metadata} that is added to the image file at the moment of capture.
This metadata is typically represented as dozens of Exchangeable Image File Format (EXIF) tags that describe the camera, its settings, and postprocessing operations that were applied to the image: e.g.,~~{\small {\tt Model}: \quotes{\tt iPhone 4s}} or {\small {\tt Focal Length}: \quotes{\tt 35.0 mm}}.  We train a joint embedding through contrastive learning that puts image patches into correspondence with camera metadata (\fig{fig:teaser}a). Our model processes the metadata with a transformer~\cite{transformer} after converting it to a language-like representation. To do this conversion, we take advantage of the fact that EXIF tags are typically stored in a human-readable (and text-based) format. We convert each tag to text, and then concatenate them together.
Our model thus closely resembles contrastive vision-and-language models, such as CLIP~\cite{CLIP}, but with EXIF-derived text in place of natural language.

We show that our model can successfully estimate camera properties solely from images, and that it provides a useful representation for a variety of image forensics and camera calibration tasks. Our approaches to these tasks {do not} require camera metadata at test time. Instead, camera properties are estimated implicitly from image content via multimodal embeddings.

We evaluate the learned feature of our model on two classification tasks that benefit from a low-level understanding of images: estimating an image's radial distortion parameter, and distinguishing real and manipulated images. We find that our features significantly outperform alternative supervised and self-supervised feature sets. 

We also show that our embeddings can be used to detect image splicing \quotes{zero shot} (i.e., without labeled data), drawing on recent work~\cite{bondi2017tampering,mayer2018learned,huh2018fighting} that detects inconsistencies in camera fingerprints hidden within image patches. Spliced images contain content from multiple real images, each potentially captured with a different camera and imaging pipeline. Thus, the embeddings that our model assigns to their patches, which convey camera properties, will have less consistency than those of real images. We detect manipulations by flagging images whose patch embeddings do not fit into a single, compact cluster. We also localize spliced regions by clustering the embeddings within an image (Fig.~\ref{fig:teaser}b).

\noindent We show through our experiments that:
\begin{itemize}[leftmargin=*,topsep=1pt, noitemsep]
\item Camera metadata provides supervision for self-supervised representation learning.
\item Image patches can be successfully associated with camera metadata via joint embeddings. %
\item Image-metadata embeddings are a useful representation for forensics and camera understanding tasks.
\item Image manipulations can be identified \quotes{zero shot} by identifying inconsistencies in patch embeddings.

\end{itemize}

\begin{figure}[t]
    \centering
    \vspace{-3mm}
    \includegraphics[width=\linewidth]{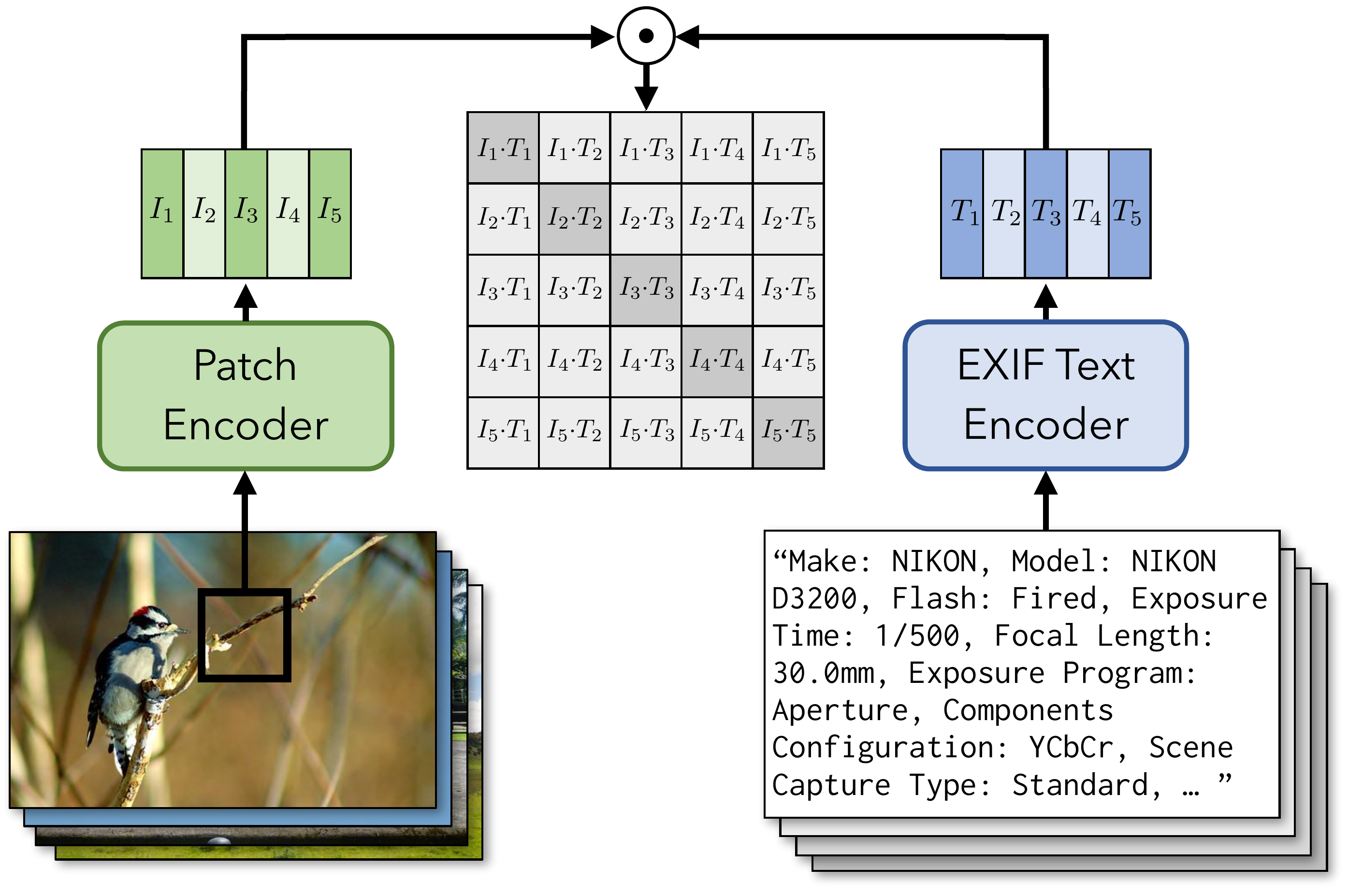}
    \vspace{-6mm}
    \caption{{\bf Cross-modal image and camera metadata model.} We use contrastive learning to associate each image patch with the EXIF metadata that was extracted from its image file. We represent the metadata as text, which is obtained by concatenating the EXIF tags together. We then process it using a transformer.} \vspace{-3mm}
    \label{fig:method}
    \vspace{-1mm}
\end{figure}

\section{Related Work}

\paragraph{Estimating camera properties.} Camera metadata has been used for a range of tasks in computer vision, such as for predicting focal length~\cite{focal_length, focall_length2, radial_cali, camera_cali}, performing white balancing~\cite{white_balance, white_balance2} and estimating camera models \cite{estCam1,estCam2,estCam3}. It has also been used as extra input for recognition tasks~\cite{estEXIF3, estEXIF1}. Instead of estimating camera properties directly (which can be highly error prone~\cite{huh2018fighting}), our model predicts an embedding that distinguishes a patch's camera properties from that of other patches in the dataset.

\mypar{Image forensics.}
Early work used physically motivated cues, such as misaligned JPEG blocks~\cite{JPEG1, JPEG2}, color filter array mismatches~\cite{cfa1, cfa2, cfa3, cfa4}, inconsistencies in noise patterns~\cite{noise1, noise2, noise3, noise4}, and compression or boundary artifacts~\cite{artifact1, artifact2, artifact3, boundary_artifact}. Other works use supervised learning methods~\cite{supervise2, supervise3, simulate1, simulate2, simulated3, mantraNet, robustNet, wang2019detecting, wang2019cnn, rossler2019faceforensics++}. The challenge of collecting large datasets of fake images has led to alternative approaches, such as synthetic examples~\cite{simulate1, simulate2, simulated3,robustNet}. Other work uses self-supervised learning, such as methods based on denoising~\cite{noiseprint}, or that detect image manipulations by identifying image content that appears to come from different camera models~\cite{CamToForensic,estCam2,mayer2018learned}. 
Huh \etal~\cite{huh2018fighting} learned a patch similarity metric in two steps: they determined which EXIF tags are shared between the patches, then use these binary predictions as features for a second classifier that predicts whether two patches come from the same (or different) images. In contrast, we obtain a visual similarity metric that is well-suited to splice localization directly from our multimodal embeddings.

\mypar{Language supervision in vision.} Recent works have obtained visual supervision from language. The formulation includes specific keyword prediction~\cite{keyword}, bag-of-word multilabel classification~\cite{bag-of-word}, $n$-gram classification \cite{n-grams} and autoregressive language models~\cite{virtex, ICMLM, VIRT}. Recently, Radford et al.~\cite{CLIP} obtained strong performance by training a contrastive model on a large image-and-language dataset. Our technical approach is similar, but uses text from camera metadata in lieu of image captions.

Work in text-to-image synthesis often exploits camera information through prompting, such as by adding text like ``DSLR photo of...'' or ``Sigma 500mm f/5'' to prompts~\cite{dalleprompt2022}. These methods, however, learn these camera associations through the (relatively rare) descriptions of cameras provided by humans, while ours learns them from an abundant and complementary learning signal, camera metadata.

\section{Associating Images with Camera Metadata}

We desire a visual representation that captures low level imaging properties, such as the settings of the camera that were used to shoot the photo. We then apply this learned representation to downstream tasks that require an understanding of camera properties.

\subsection{Learning Cross-Modal Embeddings}
  
We train a model to predict camera metadata from image content, thereby obtaining a representation that conveys camera properties. 
Following previous work in multimodal contrastive learning \cite{CLIP}, we train a joint embedding between the two modalities, allowing our model to avoid the (error prone) task of directly predicting the attributes.
Specifically, we want to jointly learn an image encoder and metadata encoder such that, given $N$ images and $N$ pieces of metadata information, the corresponding image--metadata pairs can be recognized by the model by maximizing embedding similarity.
We use full-resolution image patches rather than resized images, so that our model can analyze low-level details that may be lost during downsampling.

Given a dataset of image patches and their corresponding camera metadata $\{(\bv_i, \bm_i)\}_{i=1}^N$, we learn visual and EXIF representations $f_\theta(\bv)$ and $g_\phi(\bm)$ by jointly training $f_\theta$ and $g_\phi$ using a contrastive loss~\cite{kldivergence}:
\begin{equation}
  \mathcal{L}_{i}^{V, M} = -\log \frac{\exp{(f_\theta(\bv_i) \cdot g_\phi(\bm_i) / \tau)}}{\sum_{j=1}^N \exp(f_\theta(\bv_i) \cdot g_\phi(\bm_j)/\tau)},
  \label{eq:contrast}
\end{equation}
where $\tau$ is a small constant. Following prior work~\cite{CLIP}, we define an analogous loss $\mathcal{L}^{M, V}$ that sums over visual (rather than metadata) examples in the denominator, and minimize a combined loss $\mathcal{L} = \mathcal{L}^{V, M} + \mathcal{L}^{M, V}$.

\subsection{Representing the Camera Metadata} 

This formulation raises a natural question: how should we represent the metadata? The metadata within photos is stored as a set of EXIF tags, each indicating a different image property as shown in \tbl{tab:exif_table}. EXIF tags span a range of formats and data types, and the set of tags that are present in a given photo can be highly inconsistent. Previous works that predict camera properties from images typically extract attributes of interest from the EXIF tags, and cast them to an appropriate data format --- \eg, extracting a scalar-valued focal length category. {This tag-specific processing limits the amount of metadata information that can be used as part of learning, and requires special-purpose architectures.}%

We exploit the fact that EXIF tags are typically stored in a human-readable format and can be straightforwardly converted to text (\fig{fig:method}). This allows us to directly process camera metadata using models from natural language processing --- an approach that has successfully been applied to processing various text-like inputs other than language, such as math \cite{code_for_math} and code~\cite{code_for_transformer}. Specifically, we create a long piece of text from a photo's metadata by converting each tag's name and value to strings, and concatenating them together. We separate each tag name and value with a colon and space, and separate different tags with a space and comma. 
We evaluate a number of design decisions for this model in Sec.~\ref{sec:ablation_study}, such as the text format, choice of tags, and network architecture.

\setlength{\tabcolsep}{3pt}
\begin{table}[t!]
\begin{center}
\vspace{-2mm}
\resizebox{1.\linewidth}{!}{
{\def\arraystretch{1.2}
\begin{tabular}{@{}l l@{} l@{}}
\toprule
\textbf{EXIF tag} & \textbf{Example values} & \#values \\
\midrule
Camera Make & {\tt Canon, NIKON Corporation, Apple} & 312\\
Camera Model & {\tt NIKON D90, Canon EOS 7} & 3071\\
Software & {\tt Picasa, Adobe Photoshop, QuickTime} & 1711 \\
Exposure Time & {\tt 1/60 sec, 1/125 sec, 1/250 sec} & 2062\\
Focal Length & {\tt 18.0 mm, 50.0 mm, 6.3 mm} & 931 \\
Aperture Value & {\tt F2.8, F4, F5.6, F3.5} & 137 \\
Scene Capture Type & {\tt Landscape, Portrait, Night Scene} & 5 \\
Exposure Program & {\tt Aperture priority, Manual control} & 9\\
White Balance Mode & {\tt Auto, Manual} & 3\\
Thumbnail Compression & {\tt JPEG, Uncompressed} & 3\\
Digital Zoom Ratio & {\tt 1, 1.5, 2, 1.2} & 49 \\
ISO speed Ratings & {\tt 100, 400, 300 } & 460 \\
Shutter Speed Value & {\tt 1/60 sec, 1/63 sec, 1/124 sec} & 1161 \\
Date/Time Digitized & {\tt 2013:03:28 04:20:46} & 95932 \\
\bottomrule
\end{tabular}}}
\vspace{-2mm}
\caption{{\bf What information is contained within photo EXIF metadata?} We list several of the most common EXIF tags, along with the common values and number of values they contain in the YFCC100M dataset~\cite{yfcc100m}. }
\label{tab:exif_table} 
\vspace{-5mm}
\end{center}
\end{table}
\setlength{\tabcolsep}{6pt}

\subsection{Application: Zero-shot Image Forensics} \label{sec:application}

\begin{figure*}[t]
\vspace{-4mm}
    \centering
    \includegraphics[width=\textwidth]{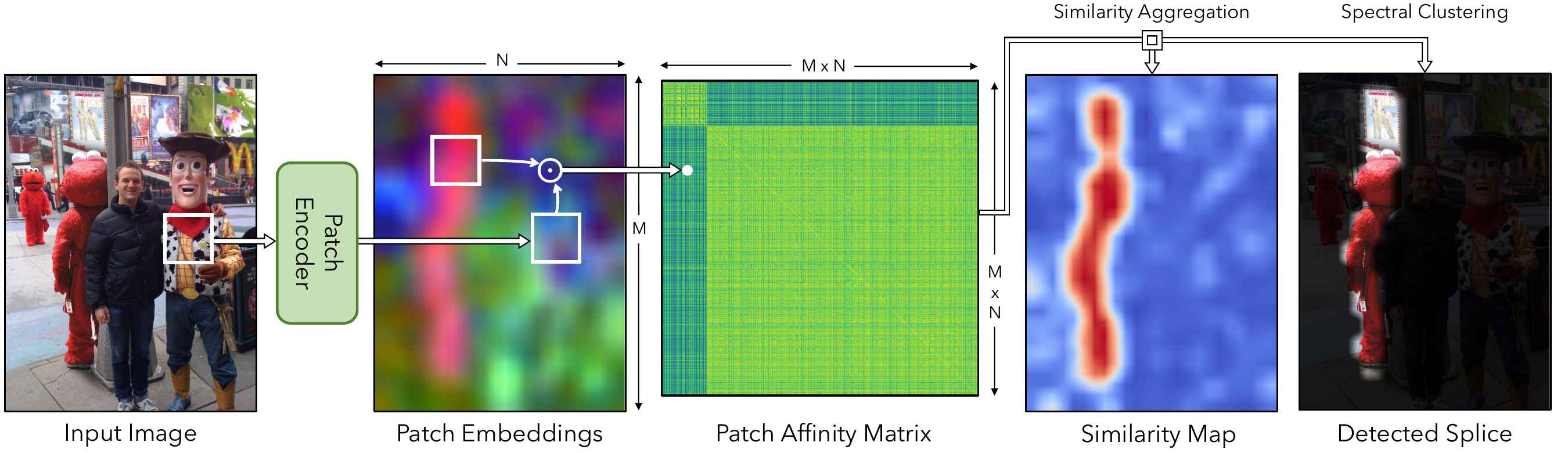}
    \vspace{-5mm}
    \caption{{\bf Zero shot splice localization}. Given a spliced image (left), we compute our cross-modal embeddings for each image patch, which we visualize here using projections onto the top 3 principal components. We then compute the affinity matrix by taking dot product for every pair of patches. We localize the spliced region by clustering these embedding vectors. } \vspace{-3mm}
    \label{fig:forensic_method}
    \vspace{-2mm}
\end{figure*}

After learning cross-modal representations from images and camera metadata, we can use them for downstream tasks that require an understanding of camera properties.  One way to do this is by using the learned visual network features as a representation for classification tasks, following other work in self-supervised representation learning~\cite{momentum,simclr}. We can also use our learned visual embeddings to perform ``zero shot'' {\em image splice detection}, by detecting inconsistencies in an input image's imputed camera properties.

Spliced images are composed of regions from multiple real images. Since they are typically designed to fool humans, forensic models need to rely more on subtle (often non-semantic) cues to detect them.
{We got inspiration from Huh~\etal~\cite{huh2018fighting}, which predicts whether two image patches share the same camera properties. If two patches are predicted to have very different camera properties, then this provides evidence that they come from different images. In our work,}
we can naturally obtain this {patch similarity} by computing the dot product between two patches' embeddings, since they have been trained to convey camera properties. We note that, unlike Huh~\etal~\cite{huh2018fighting}, we do not train a second, special-purpose classifier for this task, nor do we use augmentation to provide synthetic training examples (\eg, by applying different types of compression to the patches).

To determine whether an image is likely to contain a splice, we first compute an affinity matrix $A_{ij} = f_\theta(\bv_i) \cdot f_\theta(\bv_j)$ whose entries are the dot product between patches' normalized embedding vectors. We score an image $\bv$ using the sum of the exponentiated dot products between embeddings, $\phi(\bv) = \sum_{i,j} \exp(A_{ij}/\tau)$. This score indicates the likelihood that the image is unmodified, since high dot products indicate high similarity in imputed camera properties. To localize the spliced image regions within an image, we aggregate the similarity scores in $A_{ij}$ by clustering the rows using mean shift, following~\cite{huh2018fighting}. This results in a {\em similarity map} indicating the likelihood that each patch was extracted from the  largest source photo that was used to create the composite. Alternatively, we can visualize the spliced region by performing spectral clustering via normalized cuts~\cite{huh2018fighting,shi2000normalized}, using $A_{ij}$ as an affinity matrix between patches. We visualize this approach in \fig{fig:forensic_method}.

\section{Results}

\subsection{Implementation} 

\paragraph{Architecture.} 
We use ResNet-50 pretrained on ImageNet as our image encoder. We found that the text encoder in models trained on captioning, such as CLIP~\cite{CLIP}, were not well-suited to our task, since they 
place low limits on the number of tokens. For the EXIF text encoder, we use DistilBERT~\cite{sanh2019distilbert} pretrained on Wikipedia and the Toronto Book Corpus~\cite{zhu2015aligning}. 
We compute the feature representation of the EXIF as the activations for the end-of-sentence token from the last layer which is layer normalized and then linearly projected into multi-modal embedding space.

\mypar{Training.} To train our model, we use 1.5M full-resolution images and EXIF pairs from {a random subset of} YFCC100M~\cite{yfcc100m}. We discard images that have less than 10 of the EXIF tags. Because many images only have a small number of EXIF tags available, we only use tags  that are present in more than half of these images. This results in 44 EXIF tags (see 
 supplementary for the complete list). In contrast to other work~\cite{huh2018fighting}, we do not rebalance the images to increase the rate of rare tags. %
During training, we randomly crop $124 \times 124$ patches from high-resolution images.
We use the AdamW optimizer~\cite{adam} with a learning rate of $10^{-4}$, weight decay of $10^{-3}$, and mixed precision training. We use a cosine annealing learning rate schedule~\cite{cosine_lr}. The batch size is set to 1024, and we train our model for 50 epochs.

\mypar{Other model variations.}
To study the importance of metadata supervision on the learned representation, we train a similar model that performs contrastive learning but does not use metadata.
The model resembles image-image contrastive learning~\cite{zoran2015learning,simclr,momentum,huh2018fighting}, which has been shown to be highly effective for representation learning, and which may learn low-level camera information~\cite{context_prediction}.
Different from typical contrastive learning approaches, we use strict cropping augmentation so that the views for our model (Eq.~\ref{eq:contrast}) come from different crops of the same image, to encourage it to learn low-level image features. We call this model \textbf{CropCLR}.
Additionally, we evaluate a number of ablations of our model, including models that are trained with individual EXIF tags, that use different formats for the EXIF-derived text, and different network architectures (Table~\ref{tab:ablations}).

\setlength{\tabcolsep}{3pt}
\begin{table}[tbp]
\centering

\resizebox{1.\linewidth}{!}{
{\def\arraystretch{1.2}

\begin{tabular}{l cc cc cc} 
    \toprule
    \multirow{4}{*}{\textbf{Models}} & \multicolumn{4}{c}{\textbf{Forensics}} & \multicolumn{2}{c}{\textbf{Radial Distortion}} \\
    \cmidrule(lr){2-5}
    \cmidrule(lr){6-7}
    & 
    \multicolumn{2}{c}{CASIA I} &
    \multicolumn{2}{c}{CASIA II} &
    Dresden & RAISE \\
    
    \cmidrule(lr){2-3}
    \cmidrule(lr){4-5}
    \cmidrule(lr){6-7}
    
    & {\resize} & {\crop} & {\resize} & {\crop} & {\resize} & {\resize} \\
 
    \midrule
    
    ImageNet pretrained & 0.69 & 0.64 & 0.71 & 0.72 & 0.23 & 0.24 \\
    MoCo                & 0.67 & 0.67 & 0.68 & 0.69 & 0.24 & 0.28 \\
    CLIP                & 0.71 & 0.82 & 0.84 & 0.81 & 0.21 & 0.22 \\
    Ours - CropCLR     & 0.70 & 0.81 & 0.86 & 0.80 & 0.28 & 0.32 \\
    \baseline{Ours - Full}               & \baseline{\textbf{0.75}} & \baseline{\textbf{0.85}} & \baseline{\textbf{0.87}} & \baseline{\textbf{0.84}} & \baseline{\textbf{0.31}} & \baseline{\textbf{0.35}} \\
    \bottomrule
\end{tabular}}}
\vspace{-2mm}
\caption{We do linear probing on top of learned representation to predict two camera related properties that are not presented in EXIF files. The good performance indicates that our model learns general imaging properties. resize~ and crop~ denote the image preprocessing applied.}\vspace{-1.9mm} %
\label{feature}
\vspace{-3mm}
\end{table}
\setlength{\tabcolsep}{6pt}

\subsection{Evaluating the learned features} \label{sec: feature_eval}

First, we want to measure how well the learned features convey camera properties. 
Since EXIF file is already embedded with a lot of camera properties such as camera model, focal length, shuttle speed, etc., it should be unsurprising if we can predict those properties from images (we provide such results in Sec. \ref{sec:ablation_study}). Instead, we want to study if the feature learned by the model can be generalized to other imaging properties that are not provided in the EXIF file. Specifically, we fit a linear classifier to our learned features on two prediction tasks: radial distortion estimation and forensic feature evaluation.

We compare the features from our image encoder with several other approaches, including supervised {\bf ImageNet} pretraining~\cite{russakovsky2015imagenet}, a state-of-the-art self-supervised model {\bf MoCo}~\cite{momentum}, {\bf CLIP}~\cite{CLIP}, which obtains strong semantic representations using natural language supervision (rather than EXIF supervision), and finally the {\bf CropCLR} variation of our model. To ensure a fair comparison, the backbone architectures for all approaches are the same (ResNet-50).

\mypar{Radial distortion estimation.}  Imperfections in camera lens production often lead to radial distortion artifacts in captured images. These artifacts are often removed as part of multi-view 3D reconstruction~\cite{snavely2006photo,hartley2003multiple}, using methods that model distortion as a 4th-order polynomial of pixel position. Radial distortion is not typically specified directly by the camera metadata, and is thus often must be estimated through calibration~\cite{bouguet2004camera}, bundle adjustment~\cite{snavely2006photo}, or from monocular cues~\cite{radial_cali}.  %

We followed the evaluation setup of Lopez et al.~\cite{radial_cali}, which estimates the quadratic term of the radial distortion model, $k_1$, directly from synthetically distorted images. This term can be used to provide an estimate of radial distortion that is sufficient for many tasks~\cite{pollefeys2004visual,radial_cali}. We synthesized the $512 \times 512$ images from the Dresden Image Database~\cite{dresden} and RAISE dataset~\cite{raise} using $k_1$ parameters uniformly sampled in the range $[-0.4, 0]$. To predict $k_1$, we used a regression-by-classification approach, quantizing the values of $k_1$ into 20 bins (We also show result in regression RMSE metrics in supplementary). We extracted features from different models, and trained a {linear classifier} on this 20-way classification problem.  We provided them with a $512 \times 512$ image as input, and obtained image features from the global average pooling layer after the final convolutional layer (i.e., the penultimate layer of a typical ResNet).

\mypar{Representation learning for image forensics.}
We evaluate our model's ability to distinguish real and manipulated images. This is a task that requires a broader understanding of low level imaging properties, such as spotting unusual image statistics. We use the {\bf CASIA \uppercase\expandafter{\romannumeral1}}~\cite{casia_v1} and {\bf CASIA \uppercase\expandafter{\romannumeral2}}~\cite{casia_v2} datasets. The former contains only spliced fakes, while the latter contains a wider variety of manipulations. We again perform linear classification using the features provided by different models. We evaluate two types of preprocessing, resizing and center cropping, to test whether this low level task is sensitive to these details.

In both tasks, we found that our model's features significantly outperformed those of the other models (Table~\ref{feature}). Our method achieves much better performance than traditional representational learning methods \cite{resnet, momentum, CLIP}, perhaps because these models are encouraged to discard low-level details, while for our training task they are crucially important. Interestingly,  the variation of our model that does not use EXIF metadata, CropCLR, outperforms the supervised \cite{resnet} and self-supervised baselines \cite{momentum}, but significantly lags behind our full method. This is perhaps because it often suffices to use high-level cues (\eg color histograms and object co-occurrence) to solve CropCLR's pretext task. This suggests  metadata supervision is an important learning signal and can effectively guide our model to learn general imaging information.

\begin{table*}[tbp]
\centering

\scalebox{0.98}{

\setlength{\tabcolsep}{5pt}
\begin{tabular}{l l cc cc cc cc cc} 
    \toprule
   \multirow{2.5}{*}{\textbf{Style}}  & \multirow{2.5}{*}{\textbf{Method}} & 
    \multicolumn{2}{c}{\textbf{Columbia}~\cite{columbia}} & 
    \multicolumn{2}{c}{\textbf{DSO}~\cite{dso}} &
    \multicolumn{2}{c}{\textbf{RT}~\cite{realistic_tampering}} &
    \multicolumn{2}{c}{\textbf{\small In-the-Wild}~\cite{huh2018fighting}} &
    \multicolumn{2}{c}{\textbf{Hays}~\cite{scene_completion}}  \\
    
    \cmidrule(lr){3-4}
    \cmidrule(lr){5-6}
    \cmidrule(lr){7-8}
    \cmidrule(lr){9-10}
    \cmidrule(lr){11-12}

     &    &  {\small\pmAP} & {\small\cIoU} & {\small\pmAP} & {\small\cIoU}& {\small\pmAP}& {\small\cIoU} & {\small \pmAP} & {\small\cIoU} & {\small\pmAP} & {\small\cIoU}\\ 
    \midrule

    \multirow{3}{*}{Handcrafted} 
    & CFA~\cite{cfa}           & \textbf{0.76} & \textbf{0.75}  & 0.24          & 0.46           & \textbf{0.40} & \textbf{0.63} & 0.27          & 0.45          & 0.22          & 0.45         \\
    & DCT~\cite{dct}           & 0.43          & 0.41           & 0.32          & \textbf{0.51}  & 0.12          & 0.50          & 0.41          & 0.51          & 0.21          & \textbf{0.47} \\
    & NOI~\cite{noi}           & 0.56          & 0.47           & \textbf{0.38} & 0.50           & 0.19          & 0.50          & \textbf{0.42} & \textbf{0.52} & \textbf{0.27} & \textbf{0.47} \\
    \midrule %
    \multirow{3}{*}{Supervised} 
    & MantraNet~\cite{mantraNet}    & \textbf{0.78} & 0.88          & 0.53          & 0.78          & 0.50          & 0.54          & 0.50          & 0.63          & 0.27          & 0.56        \\
    & MAG~\cite{fantasy}           & 0.69        & 0.77       &  0.48       & 0.56         & 0.51     & 0.55   & 0.47  & 0.59   & \textbf{0.30}  &  \textbf{0.61} \\
    & OSN~\cite{robustNet}          & 0.68          & \textbf{0.90} & \textbf{0.55} & \textbf{0.85} & \textbf{0.51} & \textbf{0.81} & \textbf{0.66} & \textbf{0.88} & 0.28 & 0.57\\ 
    \midrule %
    \multirow{4}{*}{Unsupervised} 
    & Noiseprint~\cite{noiseprint}   & 0.71          & 0.83          & \textbf{0.66} & \textbf{0.90} & \textbf{0.29} & \textbf{0.80} & 0.50          & 0.78          & 0.22          & 0.53          \\
    & EXIF-SC~\cite{huh2018fighting} & 0.89          & 0.97          & 0.47          & 0.81          & 0.22          & 0.75          & 0.49          & 0.79          & 0.26          & 0.54          \\
    & Ours - CropCLR                 & 0.87          & 0.96          & 0.48          & 0.81          & 0.23          & 0.74          & 0.47          & 0.80          & 0.26          & 0.55          \\
    & \baseline{Ours - Full}                    & \baseline{\textbf{0.92}} & \baseline{\textbf{0.98}} & \baseline{0.56}          & \baseline{0.85}          & \baseline{0.23}          & \baseline{0.74}          & \baseline{\textbf{0.51}} & \baseline{\textbf{0.82}}& \baseline{\textbf{0.30}} & \baseline{\textbf{0.58}} \\
    \bottomrule
\end{tabular}
}
\vspace{-2mm}
\caption{\textbf{Zero shot splice localization.} We evaluate our model on several datasets using permutation-invariant mean average precision (\texttt{p-mAP}) over pixels and class-balanced IOU (\texttt{cIoU}) with optimal threshold selected per image. The result indicates that our model is comparable to state-of-the-art methods, although not specially optimized for this task. }\vspace{-2mm} %
\label{splice_loc}
\vspace{-1mm}
\end{table*}
\setlength{\tabcolsep}{6pt}

\begin{table}[ht]
    \centering
    \scalebox{0.95}{
    \begin{tabular}{l c c c} 
        \toprule
        \textbf{Dataset}  & 
        \textbf{Columbia \cite{columbia}} & 
        \textbf{DSO \cite{dso}} &  
        \textbf{RT \cite{realistic_tampering}} \\
        
        \midrule %
        CFA \cite{cfa}         & 0.83 & 0.49 & 0.54 \\
        DCT \cite{dct}         & 0.58 & 0.48 & 0.52 \\
        NOI \cite{noi}         & 0.73 & 0.51 & 0.52 \\
        \midrule
        EXIF-SC       & 0.98 & 0.61 & \textbf{0.55} \\
        Ours - CropCLR & 0.96 & 0.62 & 0.52 \\
        \baseline{Ours - Full} & \baseline{\textbf{0.99}} & \baseline{\textbf{0.66}} & \baseline{0.53}\\
        \bottomrule
    \end{tabular}
    } \vspace{-1mm}
    \caption{\textbf{Zero-shot splice detection:} We compare our splice detection accuracy on 3 datasets. We measure the mean average precision (mAP) of detecting whether an image has been spliced.}
    \label{tab:splice_detection}
    \vspace{-6mm}
\end{table}

\subsection{Zero Shot Splice Detection and Localization}

We evaluate our model on the task of detecting spliced images without any labeled training data. This is in contrast to Sec.~\ref{sec: feature_eval}, which used labeled data. We perform both splice detection (distinguish an image being spliced or not) and splice localization (localize spliced region within an image).

\mypar{Implementation.} %
For fair evaluation, we closely follow the approach of Huh \etal~\cite{huh2018fighting}. Given an image, we sample patches in a grid, using a stride such that the number of patches sampled along the longest image dimension is 25. %
To increase the spatial resolution of each similarity map, we average the predictions of overlapping patches. We consider the smaller of the two detected regions to be the splice.

\mypar{Evaluation.}  In splice localization task, we compare our model to a variety of forensics methods. These include traditional methods that use handcrafted features \cite{cfa, dct,noi}, supervised methods~\cite{mantraNet, fantasy, robustNet}, and self-supervised approaches \cite{huh2018fighting,noiseprint}. The datasets we use include {\bf Columbia}~\cite{columbia}, {\bf DSO} \cite{dso}, {\bf Realistic Tampering (RT)}~\cite{realistic_tampering}, {\bf In-the-Wild}~\cite{huh2018fighting} and {\bf Hays and Efros} inpainting images~\cite{scene_completion}.  Columbia and DSO  are created purely via image splicing, while Realistic Tampering contains a diverse set of manipulations. In-the-Wild is a splicing image dataset composed of internet images, which may also contain a variety of other manipulations. Hays and Efros~\cite{scene_completion} perform data-driven image inpainting. The quantitative comparison in terms of permuted-mAP ({\small\pmAP}) and class-balanced IoU ({\small\cIoU}) following \cite{huh2018fighting} are presented in \tbl{splice_loc}. 
 We also include splice image detection result in \tbl{tab:splice_detection}, where we compare our model to methods that enable splice detection.

 Our model ranks first or second place for metrics in most datasets, and obtains performance comparable to top self-supervised methods that are specially designed for this task. In particular, our model significantly outperforms the most related technique, EXIF-SC~\cite{huh2018fighting}. We note that both our method and EXIF-SC get relatively low performance on the Realistic Tampering dataset. This may be due to the fact that this dataset contains manipulations such as copy-move that we do not expect to detect (since both regions share the same camera properties). In contrast to methods based on segmentation~\cite{mantraNet, robustNet, noiseprint}, we do not aim to have spatially precise matches, and output relatively low-resolution localization maps based on large patches. Consequently, our model is not well-suited to detecting very small splices, which also commonly occur in the Realistic Tampering dataset. 

\begin{figure*}[thbp]
    \vspace{-5.0mm}    
    \centering
    \includegraphics[width=\textwidth]{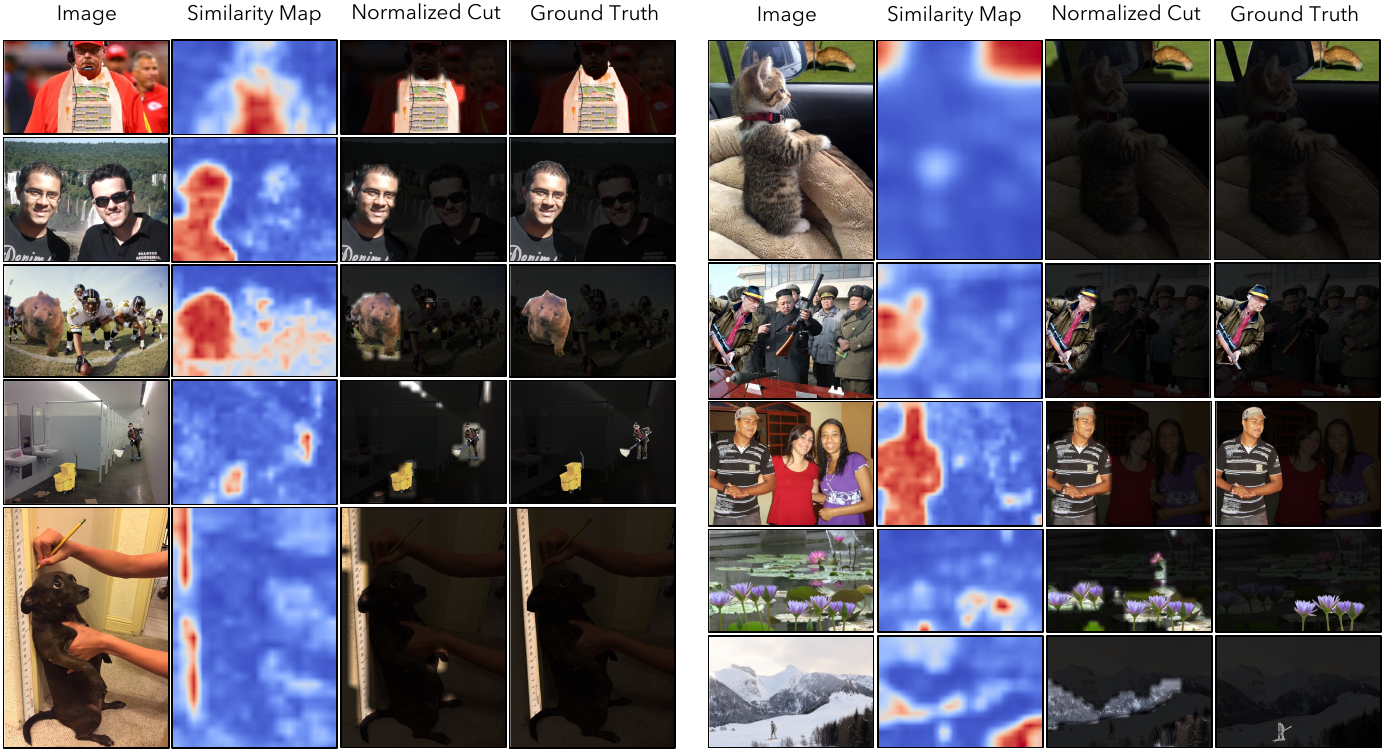}
    \vspace{-5.0mm}
    \caption{Qualitative visualization of splice localization result. We also include two typical scenarios where our model fails: copy-move tampering and very small splicing region (last two rows in right column).}
    \label{fig:my_label}
    \vspace{-2mm}
\end{figure*}

\begin{figure*}[htbp]
\vspace{-7mm}  
    \centering
    \includegraphics[width=\textwidth]{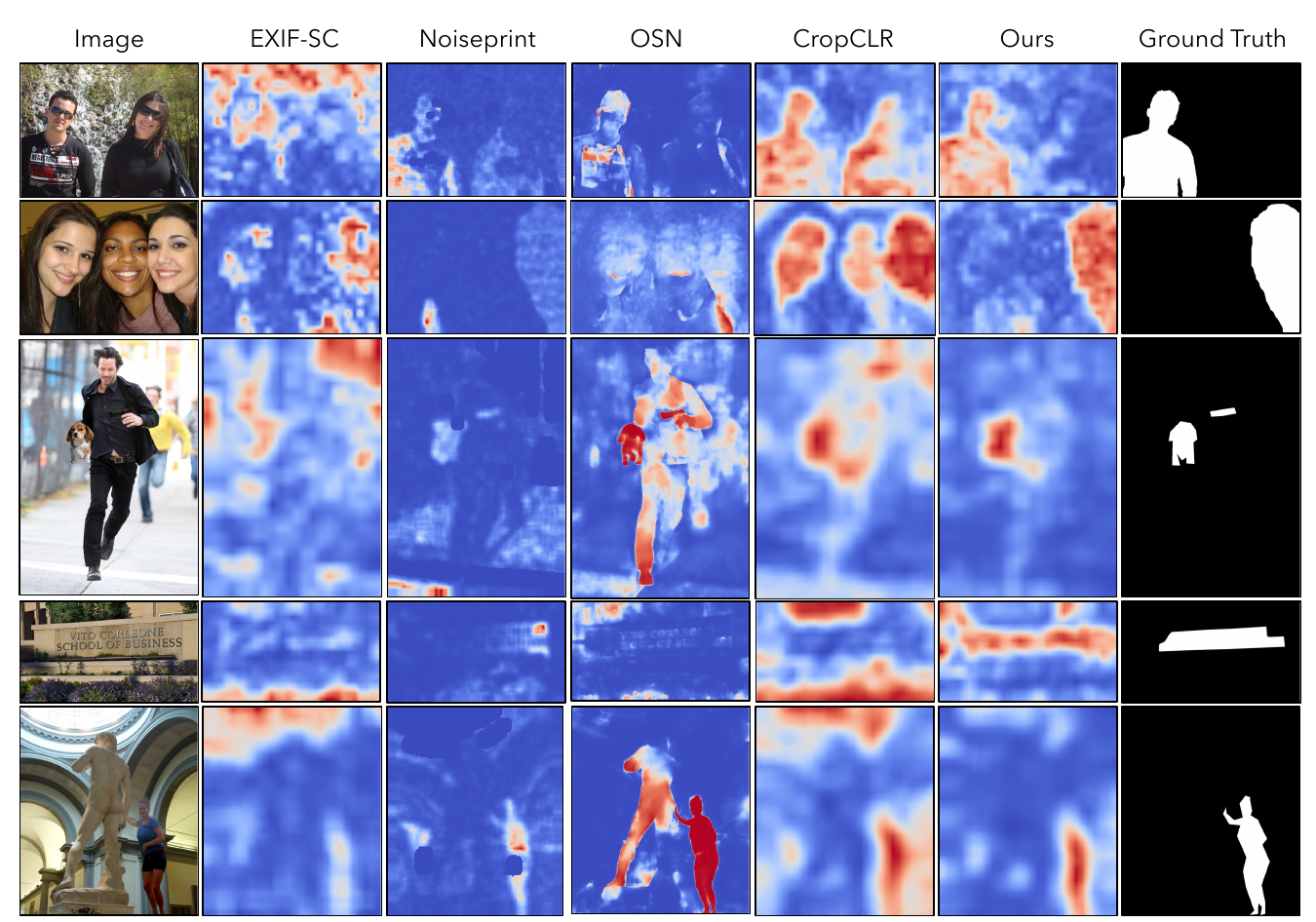}
    \vspace{-7mm}
    \caption{\textbf{Qualitative comparison to other methods.} Our method can correctly localize splices in many scenarios where other methods fail. For example, EXIF-SC~\cite{huh2018fighting} fails on overexposed image regions; OSN \cite{robustNet} and CropCLR often segment scenes based on semantics.}
     \vspace{-6mm}
    \label{fig:qual_comp}
\end{figure*}

In \fig{fig:my_label}, we show qualitative results, including both similarity maps and spectral clustering results. In \fig{fig:qual_comp}, we compare our model with those of several other techniques. Interestingly, EXIF-SC has false positives in overexposed regions (as pointed out by~\cite{huh2018fighting}), since its classifier cannot infer whether these regions are (or are not) part of the rest of the scene. In contrast, our model successfully handles these regions. CropCLR incorrectly flags regions that are semantically different from the background, because this is a strong indication that the patches come from different images. In contrast, we successfully handle these cases, since our model has no such ``shortcut'' in its learning task.

\begin{table}

\centering

\centering
\resizebox{.95\linewidth}{!}{
  {\def\arraystretch{1.1}
\begin{tabular}{clccc}
  \toprule
  & \textbf{Method} & \textbf{EXIF} & \textbf{Radial} & \textbf{Forens.} \\
  \midrule
     & Majority class baseline & 0.12 & 0.05 & 0.50 \\
  \midrule
  \multirow{5}{*}{\rotatebox[origin=c]{90}{Supervision}}   & \baseline{All EXIF tags}  & \baseline{\textbf{0.35}} & \baseline{\textbf{0.29}} & \baseline{\textbf{0.85}}\\
& CropCLR & 0.29 & 0.22 & 0.84\\
  & ``Camera Model'' tag only & 0.31 & 0.27 & 0.80\\
    & ``Color Space'' tag only & 0.12 & 0.12 & 0.61 \\
  & YFCC image descriptions & 0.15 & 0.16 & 0.70\\
  \midrule
  \multirow{4}{*}{\rotatebox[origin=c]{90}{Tag format}} & \baseline{Fixed order, w/ tag name} & \baseline{\textbf{0.35}} & \baseline{\textbf{0.29}}  & \baseline{0.85}\\
  & Fixed order, w/o tag name & \textbf{0.35} &  0.28  & \textbf{0.86}\\
  & Rand. order, w/ tag name  & 0.34 & 0.26 & 0.77\\
  & Rand. order, w/o tag name & 0.33 & 0.26 & 0.76\\
  \midrule
  \multirow{4}{*}{\rotatebox[origin=c]{90}{Architecture}} 
  &  \baseline{DistilBERT, w/ pretrained} & \baseline{0.35} & \baseline{\textbf{0.29}} & \baseline{\textbf{0.85}}\\
  & DistilBERT, w/o pretrained & \textbf{0.36} & 0.25 & 0.77\\
  & ALBERT, w/ pretrained & 0.33 & 0.29 & 0.84\\
  & ALBERT, w/o pretrained & 0.34 & 0.26 & 0.79\\
  \bottomrule
\end{tabular}}}
\vspace{-2mm}
\caption{{\bf Model ablations}. Downstream accuracy for versions of the model trained with different text supervision, representations of camera metadata, and architectures. We use linear probing to evaluate the average prediction accuracy of EXIF tag values on our YFCC test set, radial distortion estimation on Dresden dataset, and real-or-fake classification on CASIA \uppercase\expandafter{\romannumeral1} dataset. Rows with gray background (replicated for ease of comparison) represent the same model which is our \quotes{full} model.}
\label{tab:ablations} 
\vspace{-6mm}
\end{table}

\subsection{Ablation Study}\label{sec:ablation_study}

To help understand which aspects of our approach are responsible for its performance, we evaluated a variety of variations of our model, including different training supervision, representations for the camera metadata, and network architectures.

We evaluated each model's features quality using linear probing on the radial distortion estimation and splice detection task (same as Sec. \ref{sec: feature_eval}). As an additional evaluation, we classify the values of common EXIF tags by applying linear classifiers to our visual representation. We convert the values of each EXIF tag into discrete categories, by quantizing common values and removing examples that do not fit into any category. 
We average prediction accuracies over $44$ EXIF tags to obtain overall accuracy.
We provide more details in the supplementary.
All models were trained for 30 epochs on 800K images on a subset of YFCC100M dataset. The associated texts are obtained from the image descriptions and EXIF data provided by the dataset.

\begin{figure}[t]
    \centering
    \includegraphics[width=\linewidth]{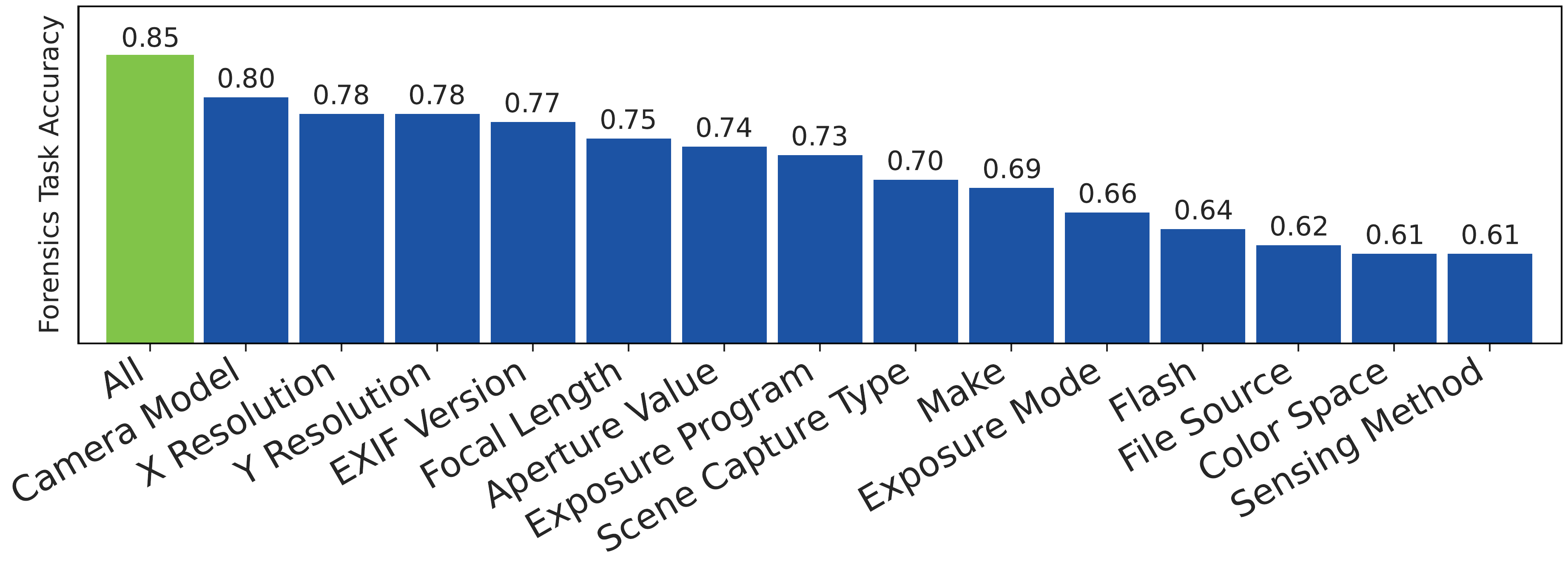}
    \vspace{-5.5mm}
    \caption{{\bf Per-tag forensics task accuracy}. We train various models supervised by individual EXIF tags, then evaluate the learned representations for splice detection task on CASIA \uppercase\expandafter{\romannumeral1}.} 
    \vspace{-5mm}
    \label{fig:barplot}
\end{figure}

\mypar{Metadata supervision.} 
We evaluate a variation of our model that trains using the image descriptions provided by YFCC100M in lieu of camera metadata, as well as models supervised by individual EXIF tags (\tbl{tab:ablations}).   For the variations supervised by a single EXIF tag,  we chose $14$ common tags for this experiment, training a separate network for each one. The results of the per-tag evaluation is shown in Fig. \ref{fig:barplot}. These results suggest that having access to the full metadata provides significantly better performance than using individual tags. Moreover, there is a wide variation in the performance of models that use different tag. This may be because the high performing tags, such as {\small \tt{Camera Model}}, convey significantly more information about the full range of camera properties than others, such as {\small \tt{Color Space}} and {\small \tt{Sensing Method}}. 
These results suggest that a model that simply uses the full range of tags can extract significantly more camera information from the metadata. We also found that the variation trained on image descriptions (rather than EXIF text) performed significantly worse than other models.

\mypar{Tag format.} 
Since EXIF does not have a natural order of tags, we ask what will happen if we randomize the EXIF tag order during training.
Table \ref{tab:ablations} shows the performance drops for all three evaluations in this case.
This may be due to the fact that the Transformer model is forced to learn meaningful positional embeddings corresponding to each EXIF tag if their order keeps on changing. We also tried removing the tag names from the camera metadata and just provide the values for those keys, \eg replacing {\small {\tt Make}: {\tt Apple}} with {\small \tt{Apple}}. 
Interestingly, this model performs on par with the model that has tag names, suggesting that the network can discern information about the tags from the values alone.

\mypar{Text encoder architecture.} To test whether performance of our model tied to a specific transformer architecture, we experimented with two different transformer models, DistilBERT~\cite{sanh2019distilbert} and ALBERT~\cite{albert}. We see that both architectures obtain similar performance on all three tasks with DistilBERT slightly outperforming ALBERT. We also test how much pretraining the text encoder helps with the performance. From \tbl{tab:ablations}, we can see pretraining improves performance on the radial distortion and forensics tasks.

\section{Discussion}

In this paper, we proposed to learn camera properties by training models to find cross-modal correspondences between images and camera metadata. To achieve this, we created a model that exploits the fact that EXIF metadata can easily be represented and processed as text. Our model achieves strong performance amongst self-supervised methods on a variety of downstream tasks that require understanding camera properties, including zero shot image forensics and radial distortion estimation.
We see our work opening several possible directions. First, it opens the possibility of creating multimodal learning systems that use camera metadata as another form of supervision, providing complementary information to high-level modalities like language and sound. Second, it opens applications that require an understanding of low level sensor information, which may benefit from our feature sets. %
 
\mypar{Limitations and Broader Impacts.} We have shown that our learned features are useful for image forensics, which has potential to reduce the spread of disinformation~\cite{farid2016photo}. The model that we will release may not be fully representative of the cameras in the wild, since it was trained only on photos available in the YFCC100M datatset~\cite{yfcc100m}.

\paragraph{Acknowledgements.}
We thank Alexei Efros for the helpful discussions. This research was developed with funding from the Defense Advanced
Research Projects Agency (DARPA) under Contract
No. HR001120C0123. The views, opinions and/or findings
expressed are those of the author and should not be interpreted
as representing the official views or policies of the Department
of Defense or the U.S. Government.

{\small
\bibliographystyle{ieee_fullname}
\bibliography{exif}
}

\appendix

\section{Additional ablations}
We also experiment with different types of patch encoders and initialized them from different pretraining models. The experimental settings are exactly the same as Sec. \ref{sec:ablation_study}. ResNet-50 outperforms ViT-B/32, while the results are not significantly affected by pretraing.

\begin{table}[h]
    \centering
    \resizebox{\linewidth}{!}{
    \begin{tabular}{clccc}
      \toprule
      & \textbf{Patch encoder} & \textbf{EXIF} & \textbf{Radial} & \textbf{Forens.} \\
      \midrule
     & ImageNet pretrained + ViT-B/32  & 0.32 & 0.27 & 0.85 \\
    & ImageNet pretrained + ResNet-50 &  \textbf{0.35} & \textbf{0.29} &  \textbf{0.85}\\
      & CLIP pretrained + ViT-B/32 & 0.31 & 0.28 & 0.84\\
        & CLIP pretrained + ResNet-50 & 0.34 & \textbf{0.29} & \textbf{0.85} \\
      \bottomrule
    \end{tabular}}
    \vspace{-2mm}
    \caption{{Additional architecture and pretraining configurations.}}
    \label{tab:aditional_ablations} 
\end{table}

\section{Sensitivity to Image Compression}
\begin{figure}[h]
    \centering
    \includegraphics[width=0.4\textwidth]{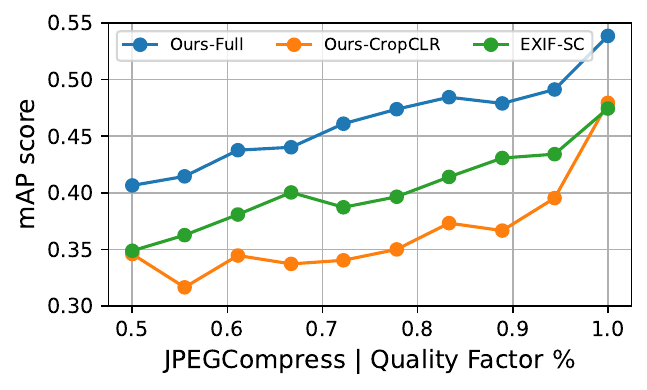}
    \caption{{\bf Effect of } Splice localization vs. JPEG compression on \textit{In the Wild}.}
    \label{fig:compression}
\end{figure}

We test the robustness of JPEG compression for different models in the zero-shot splice localization task (Fig. \ref{fig:compression}). We use \textit{in the wild} as testing dataset. All methods perform worse when noise is added which is common in forensics tasks. %

\section{Experimental Details}
We provide additional experimental details about the downstream tasks.
\subsection{Radial Distortion Model}
We follow the radial distortion model proposed in Lopez et al.~\cite{radial_cali}. Let $(x, y)$ represent the normalized image pixel coordinate. Radial distortion can be modeled as scaling the normalized coordinates by a factor of $d$, which is a function of the distance from pixel location to center of image $r$ and distortion parameters $k_1$ and $k_2$:
\vspace{-1mm}
\begin{equation}
    d=1+k_1 r^2+k_2 r^4
\end{equation}
and set $\left(x_d,y_d\right)=(d x, d y)$. %
Since the relationship between $k_1$ and $k_2$ can be approximated modeled as~\cite{radial_cali}:
\begin{equation}
    k_2=0.019 k_1+0.805 k_1^2,
\end{equation}
aim to predict only $k_1$.

To address the concern about sparsity of discrete bins we used in Table \ref{feature}, we provide another set of experiments in which we finetune and regress distortion via L2 loss. The result in terms of root mean square error (RMSE) are provided in Table \ref{tab:radial_ablations}. The finding is consistent with Table \ref{feature} that we outperform other weights.
\begin{table}[h]
    \centering
    \resizebox{0.8\linewidth}{!}{
    \begin{tabular}{clcc}
      \toprule
      & \textbf{Dataset} & \textbf{Dresden} & \textbf{RAISE} \\
      \midrule
     & ImageNet pretrained   & 0.11 & 0.12 \\
    & MoCo &  0.12 & 0.11 \\
      & CLIP & 0.10 & 0.11\\
        & Ours - CropCLR & \textbf{0.06} & 0.08  \\
        & Ours - Full & \textbf{0.06} & \textbf{0.04} \\
      \bottomrule
    \end{tabular}
    }
    \caption{\footnotesize Radial distortion regression (RMSE error)}
    \label{tab:radial_ablations} 
\end{table}

\subsection{EXIF Prediction Application}
We provide implementation details for the downstream application of predicting EXIF tags from visual features (Sec.~\ref{sec: feature_eval}). To formulate the problem as a classification task, we convert the values of each EXIF tag into discrete categories, using the following rules: if an EXIF tag has less than 20 distinct values, we use each value as a class. For example, the {\tt white balance mode} tag has only two values {\tt auto, manual}, each of which becomes a category. If an EXIF tag has continuous values (e.g., {\tt focal length}) or more than 20 discrete value (e.g., {\tt camera model}), we will quantize its common values to a set of bins using hand-chosen rules, and remove examples that do not fit into any category. For example, for the {\tt camera model} tag, which holds a sparse set of camera models, we merge their value according to their brand (value {\tt NIKON D90} will fall into {\tt NIKON} category). 
We define common values to be those that occur with probability greater than 0.1\% in the dataset.

\subsection{Linear Probing Implementation Details}
The linear probing experiment set up is as follows: We follow the approach from Chen et al.~\cite{empirical}. We use Adam optimizer with no weight decay, and set learning rate to be $0.01$ and optimizer momentum to be $\beta_1, \beta_2=0.9,0.95$. We also normalize the image features before providing them to the linear classifier. We use a batch size of 1024, and we train the classifier for 20 epochs.

\section{Confusion Plot}

\begin{figure}
    \centering
    \includegraphics[width=0.45\textwidth]{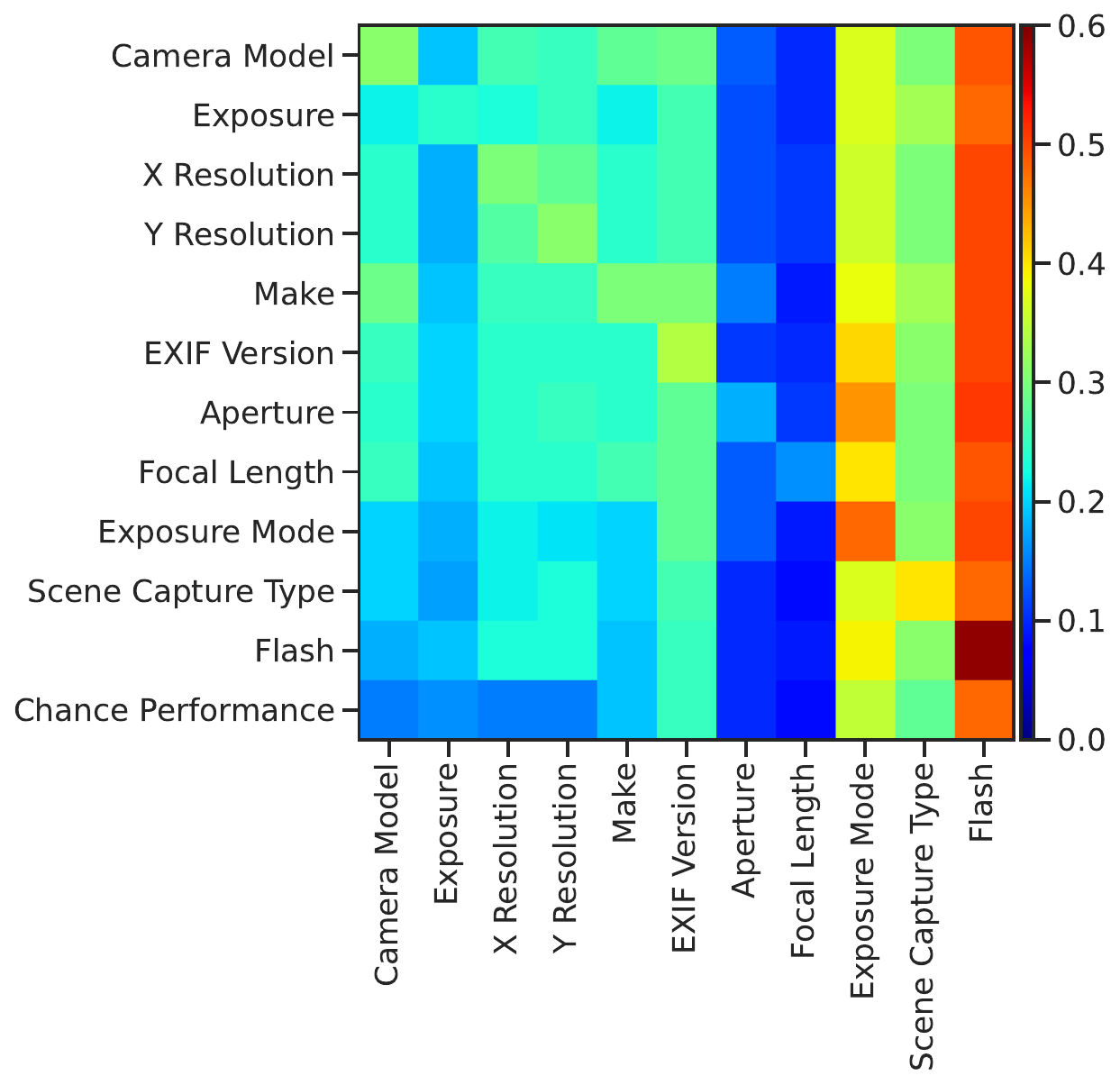}
    \vspace{-5mm}
    \caption{Confusion matrix of EXIF tag prediction accuracy. Each model is trained on one tag and tasked with predicting another tag. Each row corresponds to a model trained with a single tag, and each column represents the prediction accuracy for another tag. }
    \label{fig:confusion_plot}
\end{figure}

We ask how the performance of a model trained using a specific tag performs when it is tasked with predicting the value of other tags. This may indicate the generalizability of the training tag. We therefore take the per-tag models (same as Fig. \ref{fig:barplot}) and measure their prediction accuracies of different tag values (see Fig.~\ref{fig:confusion_plot}).  The result shows models trained on some tags may contain useful information to be generalized to other tags, such as the model trained with ``camera make'' performs well in ``camera model" and ``aperture'' predictions. In contrast, that model trained on tags that don't have rich values or information (like Flash) can not generalize to other tags well.

\section{EXIF Metadata Analysis}

In this section, we provide a detailed analysis of EXIF metadata information  in both the training and testing datasets. We hope this could help readers understand more about the characteristic of EXIF data.

\subsection{Complete list of EXIF tags used in training}
We present the complete list of EXIF tags being used by our model in Table~\ref{tab:full_exif_table}, along with representative values and the total number of values.

\subsection{Metadata in downstream tasks}
We analyze the distribution of metadata that is provided in each evaluation set, and compare it to the tags in our YFCC training set (Table \ref{tab:splice_detection}). First, we found that most of the metadata tags are not available in testing datasets because they are often removed for privacy reasons. However, the performance of our model will not be affected by this issue since it does not require metadata at test time. Second, for the tags that are available in testing datasets, nearly all the values have appeared in training time, indicating the diversity of training data. 

\begin{table}[ht]
    \vspace{-2mm}
    \centering
    \scalebox{0.6}{
    \begin{tabular}{l l l} 
        \toprule
        \textbf{Dataset}  & 
        \textbf{Available metadata and tag counts} & 
        \textbf{Values missing in training} \\
        
        \midrule %
        Dresden          & Camera Model (73), WH (12), Flash (2) & All included (0) \\
        RAISE         & Camera Model (3), WH (3), Color Space (2) & ``Color Space: \textit{Adobe RGB}'' (1) \\
        CASIA \uppercase\expandafter{\romannumeral1} \& \uppercase\expandafter{\romannumeral2}         & All missing & -- \\
        DSO       & WH (5) &  All included (0) \\
        Columbia       & Camera Model (4), WH (2) & All included (0)  \\
        RT & Camera Model (4), WH (1)   & All included (0)\\
        Hays & Camera Model (4), WH (3) & All included (0)\\
        In the Wild & WH (64) & Various WH  (17)\\
        \bottomrule
    \end{tabular}
    }
    \vspace{-2mm}
    \caption{\footnotesize Downstream metadata distribution (WH = Width/Height tags, distinct values per tag in parentheses). }
    \label{tab:splice_detection}
    \vspace{-5mm}
\end{table}

\setlength{\tabcolsep}{3pt}
\begin{table}[t]
\begin{center}
\vspace{-2mm}
\resizebox{1.0\linewidth}{!}{
{\def\arraystretch{1.2}
\begin{tabular}{lll}
\toprule
\textbf{EXIF tag} & \textbf{Example values}& \textbf{\# values}\\
\midrule

Aperture Value & {\tt F2.8, F4, F5.6, F3.5} & 137\\
Camera Make & {\tt Canon, NIKON Corporation, Apple} & 312\\
Camera Model &  {\tt NIKON D90, Canon EOS 7} & 3071 \\
Color Space & {\tt sRGB, Undefined} & 3 \\
Components Configuration & {\tt YCbCr, CrCbY} & 10 \\
Compressed Bits & {\tt 4 bits per pixel} & 6\\
Custom Rendered & {\tt Custom process, Normal process} & 3 \\
Data/Time & {\tt 2013:03:28 04:20:46} &  95982\\
Data/Time Digitized & {\tt 2013:03:28 04:20:46} & 95932\\
Data/Time Original & {\tt 2013:03:28 04:20:46} & 95839\\
Digital Zoom Ratio & {\tt 1, 1.5, 2, 1.2} & 49\\
Exif Image Height & {\tt 2592 pixels, 2304 pixels} & 3325 \\
Exif Image Width & {\tt 2592 pixels, 2408 pixels} & 3787\\
Exif Version & {\tt 2.21, 2.20, 2.30} & 14\\
Exposure Bias Value & {\tt 0 EV, -1 EV, 1 EV} & 71\\
Exposure Mode & {\tt Auto exposure, Manual exposure} & 4 \\
Exposure Program & {\tt Aperture priority, Manual control} & 9\\
Exposure Time & {\tt 1/60 sec, 1/125 sec, 1/250 sec} & 2062 \\
F-Number & {\tt F2.8, F5.6, F4} & 150 \\
File Source & {\tt Digital Still Camera, Print Scanner} & 6 \\
Flash & {\tt Unfired, Fired(red-eye reduction)} & 20 \\
FlashPix Version & {\tt 1.00, 0.10, 1.01} & 14 \\
Focal Length & {\tt 18.0 mm, 50.0 mm, 6.3 mm} & 931\\
Focal Place X Resolution & {\tt 292 dots per inch } & 61 \\
Focal Place Y Resolution & {\tt 292 dots per inch } & 60\\
Gain Control & {\tt Low, High} & 2 \\
ISO Speed Ratings & {\tt 100, 400, 300 } & 460\\
Interoperability Index & {\tt 0, unknown} & 2\\
Interoperability Version & {\tt 1.00, 1.10, 30.00} & 16 \\
Max Aperture Value & {\tt F2.8, F3.5, F4} & 81 \\
Metering Mode & {\tt Multi-segment, Spot, average} & 8\\
Orientation & {\tt Top, right side (Mirror horizontal)} & 2 \\
Resolution Unit & {\tt Inch, cm} & 3 \\
Scene Capture Type & {\tt Landscape, Portrait, Night Scene} & 5\\
Sensing Method & {\tt One-Chip color area sensor} & 4 \\
Shutter Speed Value & {\tt 1/60 sec, 1/63 sec, 1/124 sec} & 1161\\
Software & {\tt Picasa, Adobe Photoshop, QuickTime} & 1711  \\
Thumbnail Compression & {\tt JPEG, Uncompressed} & 3 \\
Thumbnail Length & {\tt 0 bytes, 16712 bytes} & 17743 \\
Thumbnail Offset & {\tt 5108 bytes, 9716 bytes} & 11298 \\
White Balance Mode & {\tt Auto, Manual} & 3\\
X Resolution & {\tt 72 dots per inch} & 62 \\
Y Resolution & {\tt 72 dots per inch} & 64 \\
YCbCr Positioning & {\tt datum point, Center of pixel array} & 3 \\

\bottomrule
\end{tabular}}}
\vspace{-2mm}
\caption{{\bf Full list of EXIF tags being used in training. } This extends the list from Table 1 in the main paper.}
\label{tab:full_exif_table} 
\vspace{-9mm}
\end{center}
\end{table}
\setlength{\tabcolsep}{6pt}

\section{Additional Qualitative Results}

\begin{figure}
    \centering
    \includegraphics[width=0.5\textwidth]{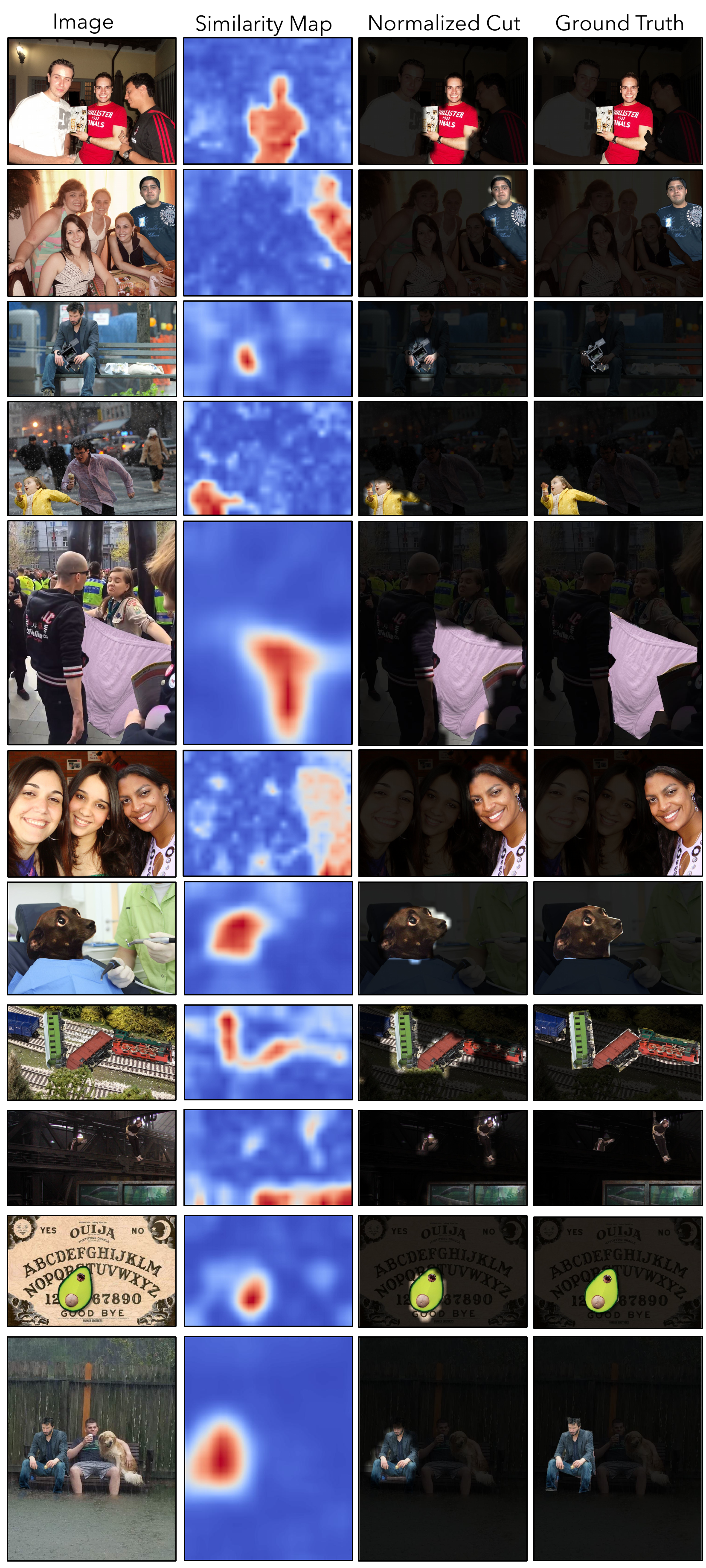}
    \vspace{-6mm}
    \caption{{Additional zero-shot splice localization results.}}
    \label{fig:qual1}
\end{figure}

\begin{figure*}[t]
    \centering
    \includegraphics[width=\textwidth]{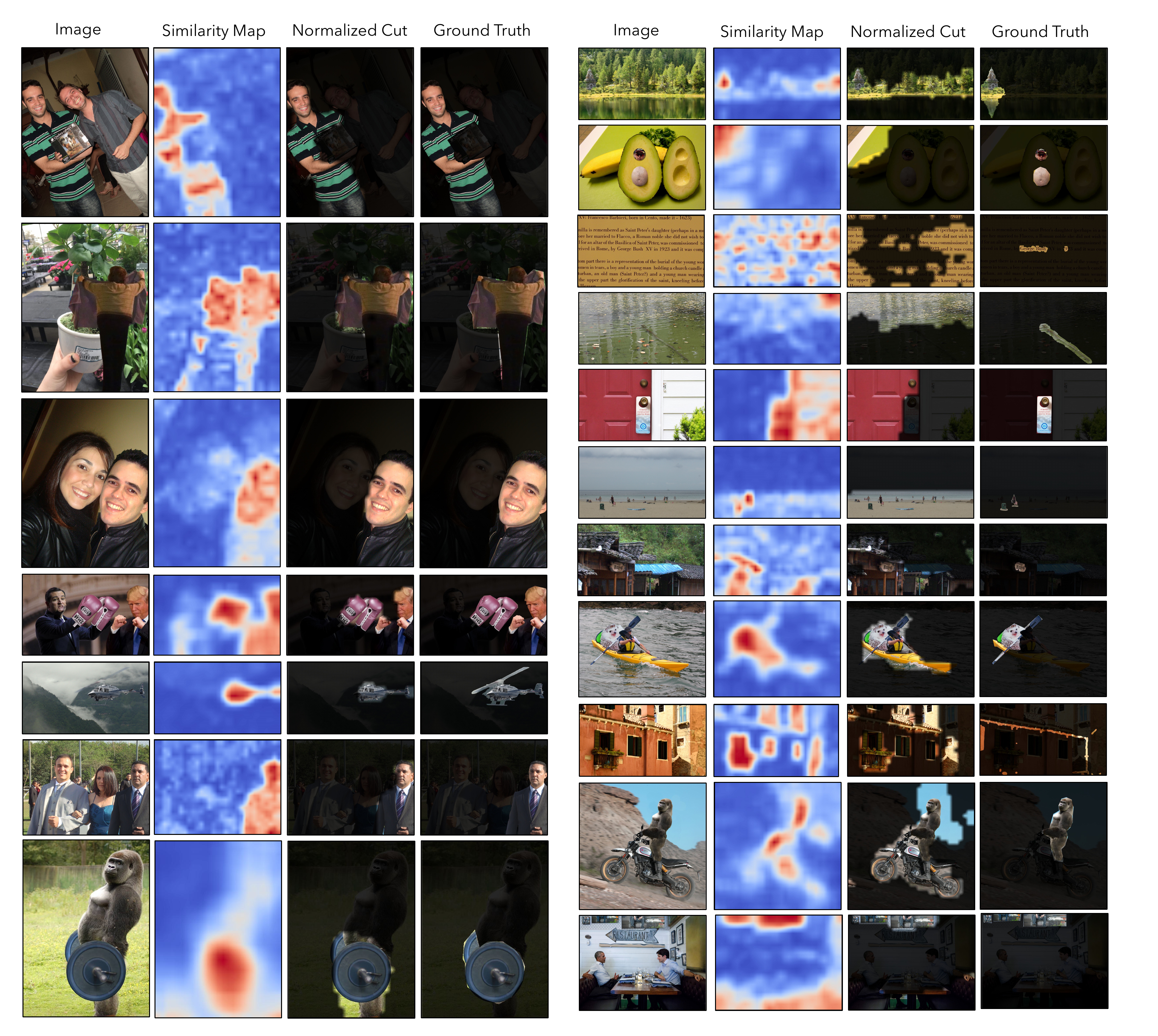}
    \vspace{-6mm}
    \caption{Additional zero-shot splice localization results: success cases (left) and failure cases (right).}
    \label{fig:qual2}
\end{figure*}

We provide additional qualitative results for our zero-shot splice localization model. In~Fig.~\ref{fig:qual1} and Fig.~\ref{fig:qual2} (left), we show accurate predictions. In Fig.~\ref{fig:qual2} (right), we show failure cases.

\end{document}